\documentclass[10pt,twocolumn,letterpaper]{article}
\usepackage{array}
\usepackage{multirow}
\usepackage{algorithm}
\usepackage{algorithmic}
\usepackage{iccv} 

\def\name{FedVLA}

\newcommand{\bfsection}[1]{\vspace*{0.0mm}\noindent\textbf{#1.}}
\usepackage{xcolor}

\definecolor{iccvblue}{rgb}{0.21,0.49,0.74}
\usepackage[pagebackref,breaklinks,colorlinks,allcolors=iccvblue]{hyperref}

\usepackage{amsmath,amsfonts,bm}

\def\eqref#1{equation~\ref{#1}}

\def\1{\bm{1}}

\DeclareMathAlphabet{\mathsfit}{\encodingdefault}{\sfdefault}{m}{sl}
\SetMathAlphabet{\mathsfit}{bold}{\encodingdefault}{\sfdefault}{bx}{n}

\def\sC{{\mathbb{C}}}
\def\sD{{\mathbb{D}}}

\def\sT{{\mathbb{T}}}

\def\paperID{6236} 
\def\confName{ICCV}
\def\confYear{2025}

\title{FedVLA: Federated Vision-Language-Action Learning with Dual Gating Mixture-of-Experts for Robotic Manipulation}

\author{
Cui Miao$^{1}$ \quad 
Tao Chang$^{1}$ \quad 
Meihan Wu$^{1}$ \quad 
Hongbin Xu$^{2}$ \quad \\
Chun Li$^{3}$ \quad 
Ming Li$^{4}$\thanks{Corresponding author} \quad 
Xiaodong Wang$^{1}$ \\
$^1$
National University of Defense Technology  \\
$^2$Bytedance Seed \quad 
$^3$Shenzhen MSU-BIT University \\
$^4$Guangdong Laboratory of Artificial Intelligence and Digital Economy (SZ)
}

\begin{document}
\maketitle
\begin{abstract}
Vision-language-action (VLA) models have significantly advanced robotic manipulation by enabling robots to interpret language instructions for task execution. However, training these models often relies on large-scale user-specific data, raising concerns about privacy and security, which in turn limits their broader adoption. To address this, we propose \name{}, the first federated VLA learning framework, enabling distributed model training that preserves data privacy without compromising performance.
Our framework integrates task-aware representation learning, adaptive expert selection, and expert-driven federated aggregation, enabling efficient and privacy-preserving training of VLA models. 
Specifically, we introduce an Instruction-Oriented Scene-Parsing mechanism, which decomposes and enhances object-level features based on task instructions, improving contextual understanding. To effectively learn diverse task patterns, we design a Dual Gating Mixture-of-Experts (DGMoE) mechanism, where not only input tokens but also self-aware experts adaptively decide their activation. Finally, we propose an Expert-Driven Aggregation strategy at the federated server, where model aggregation is guided by activated experts, ensuring effective cross-client knowledge transfer.
Extensive simulations and real-world robotic experiments demonstrate the effectiveness of our proposals. Notably, DGMoE significantly improves computational efficiency compared to its vanilla counterpart, while FedVLA achieves task success rates comparable to centralized training, effectively preserving data privacy.

\end{abstract}  
\section{Introduction}
\label{sec:intro}


Vision-language-action (VLA) models, which integrate visual perception, linguistic understanding, and robotic control, have significantly enhanced robotic manipulation by enabling robots to comprehend and execute complex tasks through natural language instructions~\cite{brohan2023rt,kim24openvla,wang2025scaling}. However, training these models requires extensive indoor datasets that capture diverse, user-specific environments, raising serious concerns about data privacy. While recent advancements have improved the capability of VLA models in robotic systems, the issue of privacy leakage remains largely unexplored~\cite{liu2024aligning}.

\begin{figure}[!t]
\centering
    \subfloat[Centralized VLA learning \label{subfig1:a}]
    {\includegraphics[width = 0.2\textwidth]{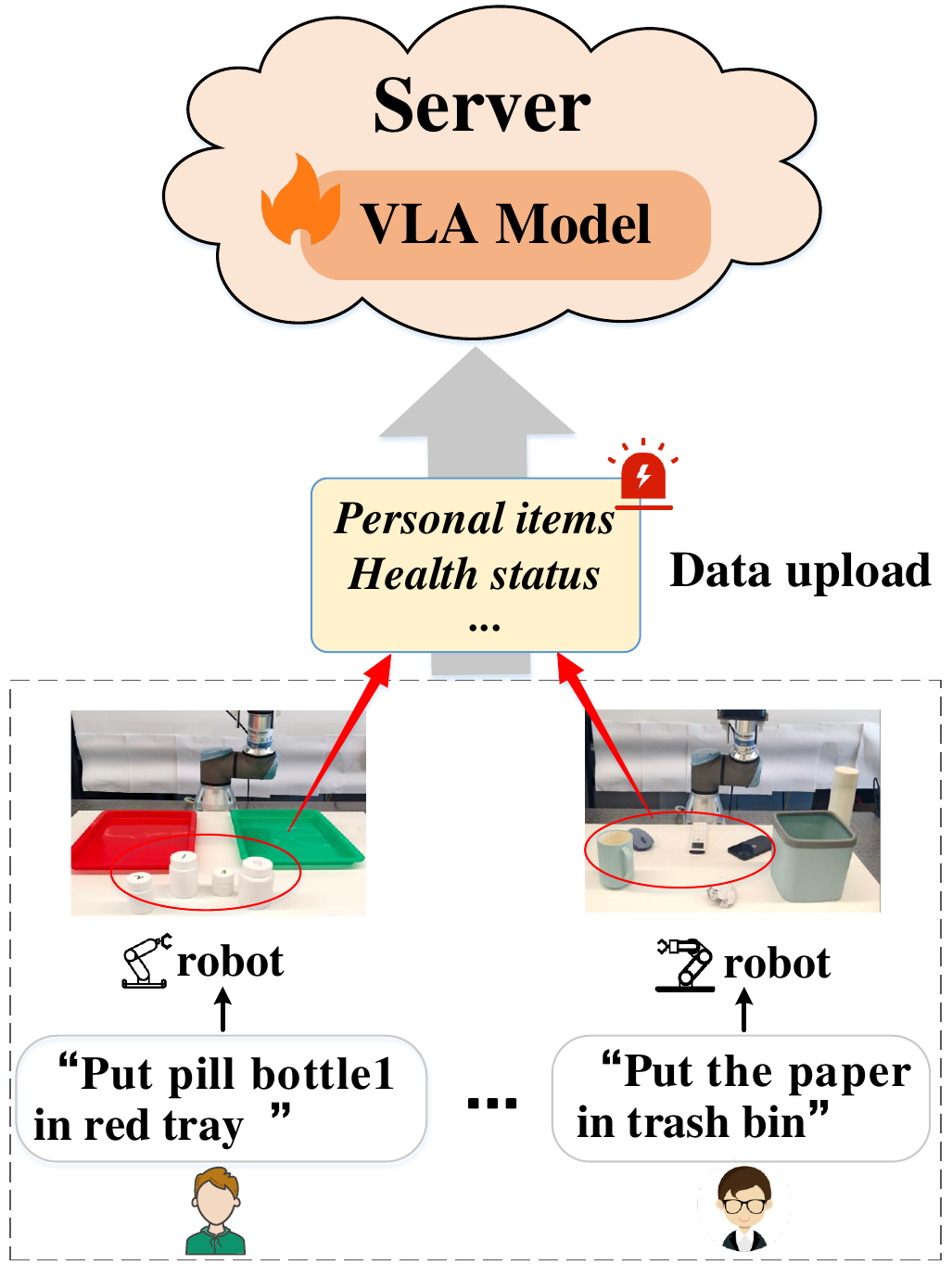}} \quad
    \subfloat[Federated VLA learning \label{subfig1:b}]
    {\includegraphics[width = 0.22\textwidth]{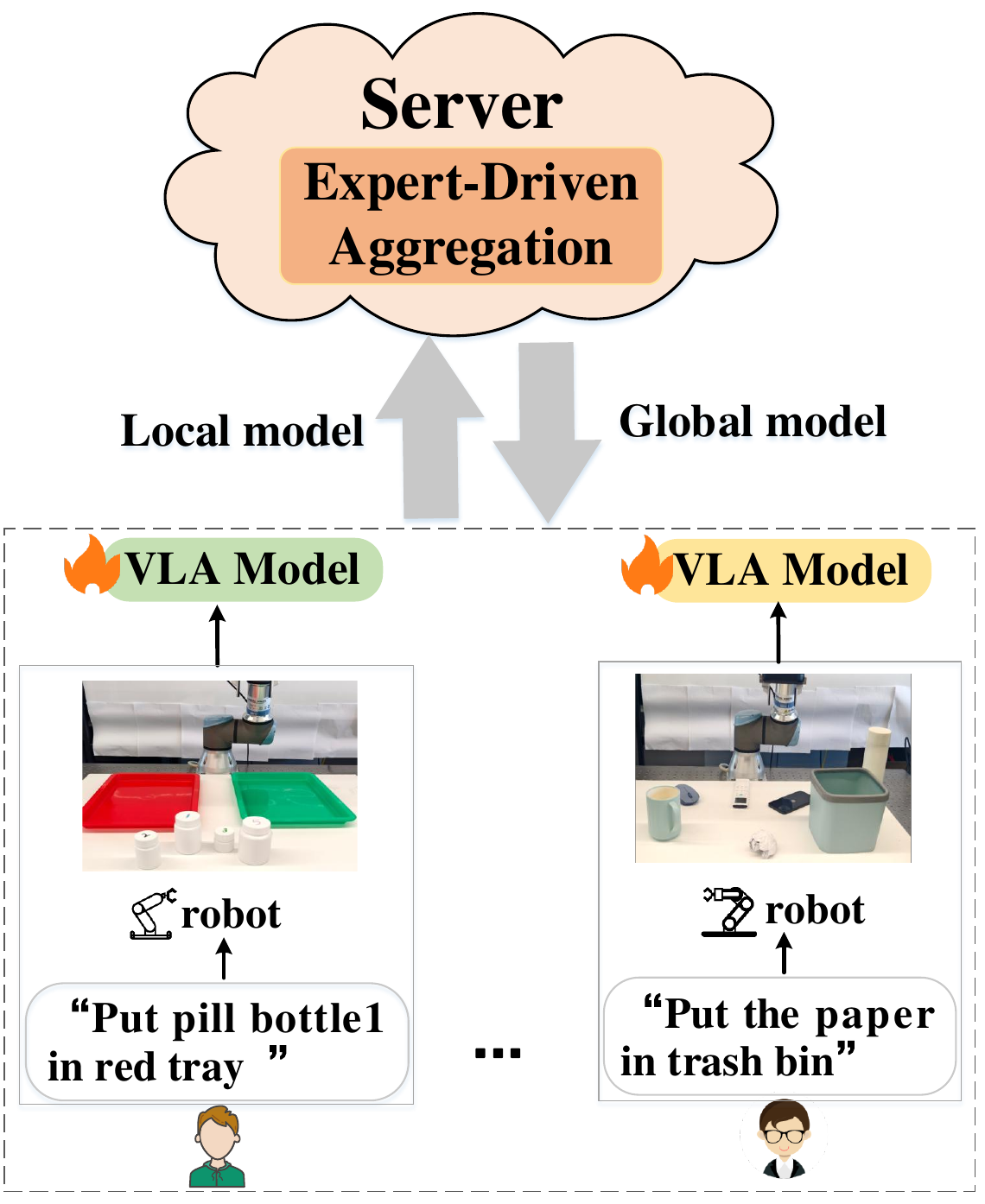}}
\caption{Comparison between centralized and federated VLA training. Centralized training collects all user data on the cloud, raising privacy concerns due to potential exposure of sensitive information. Our federated VLA framework enables decentralized training on user devices, preserving privacy while utilizing expert-driven aggregation to enhance model generalization across diverse tasks.}
\label{fig1:moti}
\end{figure}

Federated learning (FL) provides an effective privacy-preserving solution for training VLA models~\cite{mcmahan2017communication}. Unlike traditional centralized training, which requires aggregating all user data on a central server, FL enables distributed model training across multiple clients without transferring raw data. As illustrated in Figure~\ref{subfig1:a}, centralized VLA training collects and processes user data in the cloud, posing risks of privacy leakage, such as exposure of personal preferences and sensitive health conditions. In contrast, figure~\ref{subfig1:b} presents a federated VLA learning paradigm, where local models are trained on user devices, and only model updates are transmitted to the central server. This decentralized paradigm mitigates privacy risks while allowing collaborative model optimization.

Existing federated learning approaches are not well-suited for VLA learning, where robotic manipulation tasks exhibit substantial task heterogeneity across different clients. Typical FL methods, such as FedAvg \cite{mcmahan2017communication}, naively aggregate models across tasks, considering only shared synergies while overlooking their fundamental distinctions. Although recent works attempt to address this issue through merge-and-split training \cite{zhuang2023mas} or decoupled model aggregation~\cite{lu2024fedhca2}, they remain limited to single-modal settings, assuming a uniform input modality across tasks. In contrast, VLA models operate in multi-modal environments, requiring the joint processing of visual observations, language instructions, and robotic actions, which significantly increases the complexity of federated training.

Additionally,  mixture of experts (MoE) has shown promise in multi-task learning by assigning specialized experts to different tasks~\cite{jiang2024automatic,xu2024mome,chen2023mod}. However, existing methods typically select a fixed number of experts, lacking adaptability to tasks of varying complexity. This rigidity makes them inefficient for resource-constrained clients, where optimizing inference speed is critical for real-time robotic responses. These limitations highlight the need for a task-adaptive and flexible FL framework, specifically designed for multi-modal robotic learning.


In this paper, we introduce \name{}, a novel federated vision-language-action (VLA) learning framework with Dual Gating MoE, designed to preserve privacy while maintaining high task performance, as illustrated in Figure~\ref{subfig1:b}. To address task heterogeneity and enable effective knowledge aggregation in federated settings, we propose three key components.
First, we introduce an Instruction-Oriented Scene-Parsing (IOSP) module to enhance task-aware feature extraction in decentralized environments, where raw images across clients exhibit diverse backgrounds and object distributions. IOSP decomposes observation images into object-level representations guided by task instructions and leverages vision-language alignment techniques to improve contextual understanding.
Second, to enhance computational efficiency in federated learning, we develop a Dual Gating MoE (DGMoE) mechanism that enables adaptive knowledge routing. Unlike conventional MoE, which assigns tokens to a fixed number of experts, DGMoE introduces self-aware experts that can accept or reject token assignments, enabling a bidirectional selection process. This dual gating mechanism improves efficiency while preserving task performance.
Lastly, to effectively aggregate similar knowledge from diverse client models, we propose the Expert-Driven Aggregation (EDA) strategy, which dynamically assigns aggregation weights based on expert similarity across clients. By prioritizing updates from clients that activate similar experts, EDA ensures that semantically aligned knowledge is more prominently integrated at the server side, resulting in a more coherent and specialized global model.


Our main contributions in this work can be summarized as follows:
\begin{itemize}
\item We propose \name{}, the first privacy-preserving federated learning framework for VLA training, ensuring that user data remains private and never shared externally. Extensive experiments in both simulation and real-world environments demonstrate that \name{} achieves performance comparable to centralized training while preserving data privacy.

\item We introduce the Dual Gating Mixture-of-Experts, where self-aware experts autonomously determine their participation, enabling bidirectional token selection to optimize computational efficiency while maintaining task performance.

\item We propose the Expert-Driven Aggregation strategy, which dynamically assigns aggregation weights based on expert similarity across clients, ensuring that semantically aligned knowledge is more effectively integrated, thereby enhancing generalization across heterogeneous tasks.
\end{itemize}






\section{Related Works}
\label{sec:rela}
\begin{figure*}[!t]
    \centering
    \includegraphics[width=\textwidth]{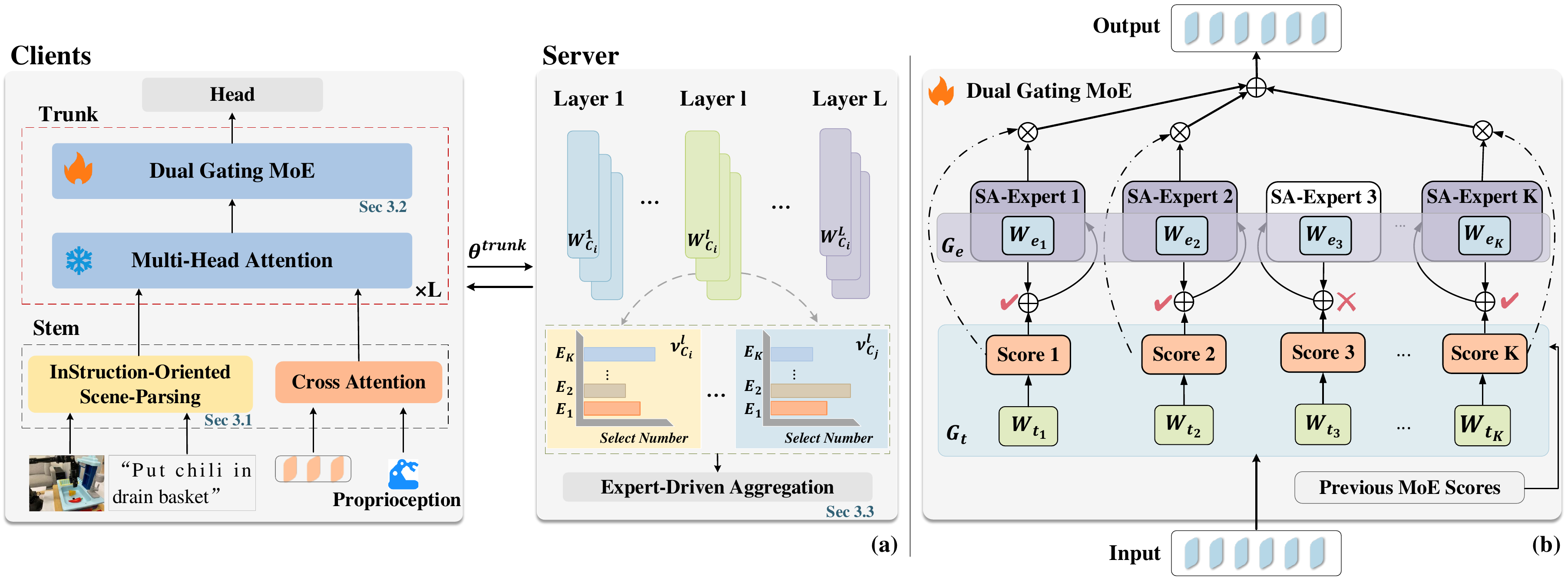}
    \caption{\textbf{Illustration of the proposed FedVLA framework and its key component.} 
(a) An overview of the FedVLA, which consists of multiple clients and a server. Each client processes vision-language-action inputs through the Stem, Trunk, and Head modules. Clients send their updated trunk parameters to the server, where Expert-Driven Aggregation selectively integrates expert-specific updates on server. The global trunk module is then distributed to clients for the next training round.
(b) Details of the proposed Dual Gating MoE  module. The input tokens first pass through the token-level gate $G_t$, computing the gating scores for each expert with residual scores from the previous MoE layer. Only the computed scores surpass the expert-level gate $G_e$, activating the corresponding expert to process the token. Finally, the expert outputs are combined to generate the final output tokens.
}
    \label{fig:overview}
\end{figure*}

\subsection{Vision-Language-Action Models}
Vision-Language Models (VLMs), which are pre-trained on large-scale datasets, have demonstrated remarkable capabilities in understanding and generating multimodal data~\cite{liu2024visual,liu2024world,team2023gemini}.Inspired by VLMs, Vision-Language-Action (VLA) models extend their capabilities by integrating robotic action prediction. For instance, RoboFlamingo~\cite{li2023vision}, RT-2~\cite{brohan2023rt}, and OpenVLA~\cite{kim24openvla} are all built upon VLMs such as OpenFlamingo~\cite{awadalla2023openflamingo}, PaLI-X~\cite{chen2023pali}, and Prismatic~\cite{karamcheti2024prismatic}, respectively. Recent work has begun exploring the training of generalist robots~\cite{wang2025scaling}, followed by fine-tuning on diverse robot datasets, enabling generalization to novel instructions, unseen objects, and distribution shifts~\cite{zawalski2024robotic}. This progress relies heavily on the availability of large-scale public datasets, such as Open-X Embodiment dataset~\cite{o2024open}. During the fine-tuning phase, samples from real-world scenarios are required, which are often generated by human activities and distributed across users' home environments. However, these samples frequently expose privacy-sensitive information.

\subsection{Privacy-Preserving Training}
Traditional centralized learning methods require uploading data from multiple sources, which poses significant privacy risks. To mitigate these concerns, researchers have proposed various privacy-preserving techniques, including federated learning (FL)~\cite{mcmahan2017communication}, differential privacy~\cite{dwork2006differential}, secure multi-party computation~\cite{goldreich1998secure}, and homomorphic encryption~\cite{albrecht2021homomorphic}. However, Most privacy-preserving training techniques have been developed for single-modality tasks, and seldomly study the data  privacy of embodied AI~\cite{zhou2022fedvln}.  In robotic manipulation, the models process multimodal sensor data, including images, language instructions, and action sequences, which can contain sensitive user interaction data. Unlike static datasets, robotic systems continuously collect and process dynamic, task-specific information, making privacy protection more complex. In this context, FL as a solution by allowing decentralized model training without sharing raw data. However, FL for robotic manipulation remains challenging due to task heterogeneity and the need for efficient model aggregation across clients.

\subsection{Mixture of Experts in Large Models}
Mixture of experts~\cite{jacobs1991adaptive,blessing2023information} is an architecture where multiple experts are trained collaboratively, with each expert responsible for a specific portion of the input space. An MoE consists of two components: a routing mechanism and a set of experts. The experts can either be part of a feedforward network (FFN) or the entire FFN layer, and the routing mechanism allows each token to select experts for processing. By activating a specific number of experts, MoE can reduce resource consumption and improve inference speed. This feature makes MoE particularly effective when integrated into large models, as seen in models like LLaVA-MoE~\cite{lin2024moe}, DeepSeekMoE~\cite{dai2024deepseekmoe}, and GPT-4~\cite{achiam2023gpt}. 
In this work, we apply MoE to a generalist robot VLA model and introduce a self-aware variant that can adaptively decide whether to be selected, significantly improving deployment efficiency.


\section{FedVLA}
\label{sec:method}


We aim to train a vision-language-action model in a federated learning (FL) framework, ensuring privacy preservation while enabling task generalization across clients.
Formally, given $N$ clients $\sC = \{  C_1, C_2, ..., C_N \}$, $N$ local tasks $\sT = \{  T_1, T_2, ..., T_N \}$, and $N$ local datasets $\sD = \{  \sD_1, \sD_2, ..., \sD_N \}$, each client $C_i$ is associated with a local task $T_i$ and a corresponding dataset $\sD_i$.
The dataset $\sD_i$ contains visual perception image $o_i$, language instruction $l_i$ of the task, proprioception $s_i$ of the robot, and action sequence $a_i$ executed by the robot:
\vspace{-2mm}
\begin{equation}
    \sD_i = \{ (o_i, l_i, s_i, a_i)_j | j=1,2,...,M \},
    \vspace{-1mm}
\end{equation}

where $M$ is the number of samples in dataset $\sD_i$.
To train a federated VLA model $\theta_G$, the overall federated optimization objective can be formulated as:
\vspace{-2mm}
\begin{equation}
    \min_{\theta_G} \sum_{i=1}^{N} w_i \sum_{(o_i, l_i, a_i,s_i) \in \sD_i} \mathcal{L}_i (f_i(\theta_G; o_i, l_i,s_i), a_i),
    \vspace{-1mm}
\end{equation}

where $\mathcal{L}_i$ is the Huber loss function computed over each client model's local dataset $\sD_i$, $f_i$ represents the local client model’s predicted action sequence, and $w_i$ denotes the weight assigned to each client during aggregation, which is further detailed in Section \ref{sec:method:EOA}.

The overview of our \name{} framework is shown in Figure \ref{fig:overview}.
The clients train their models locally and communicate with the server per round until the model converges.
The following sections introduce our key proposals: 
(1) Instruction-Oriented Scene-Parsing (Section \ref{sec:method:SPOR}) that decomposes the observation image into object-level embeddings guided by task instructions.
(2) Dual Gating Mixture-of-Experts (Section \ref{sec:method:DG-MoE}) that can adaptively select a varying number of experts for tokens based on the relevance to the specific task.
(3) Expert-Driven Aggregation (Section \ref{sec:method:EOA}) that dynamically aggregates the weights based on expert similarities across different clients in the server.

\subsection{Instruction-Oriented Scene-Parsing}
\label{sec:method:SPOR}

\begin{figure}[!t]
    \centering
    \includegraphics[width=0.48\textwidth]{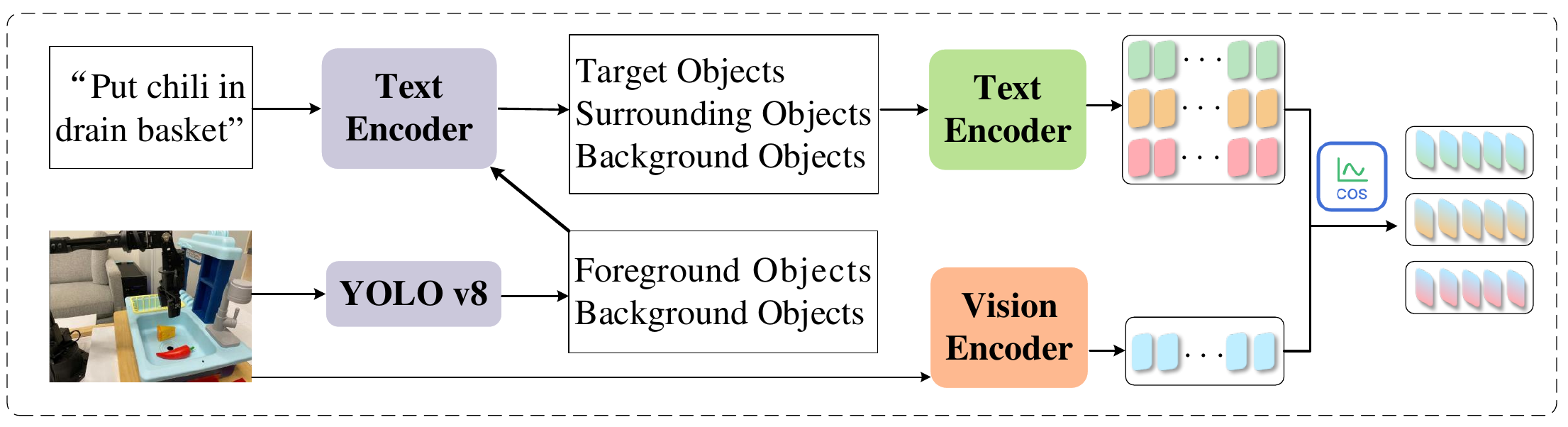}
    \caption{Instruction-Oriented Scene-Parsing module, which decomposes observations into object-level representations guided by task instructions, enhancing task-aware feature extraction through vision-language alignment.}
    \label{fig:sence}
\end{figure}


To ensure the robot focuses on task-relevant information while maintaining contextual awareness to avoid collisions, we decouple objects in the scene based on the specific robotic task. To achieve this, we propose an Instruction-Oriented Scene-Parsing (IOSP) module to represent objects within the task scene.

As illustrated in Figure \ref{fig:sence}, given an instruction and an image, we first extract Target Objects (TOs) from the instruction using named entity recognition \cite{nadeau2007survey}. Simultaneously, we employ YOLOv8 \cite{varghese2024yolov8} to detect and classify Foreground Objects (FOs) and Background Objects (BOs) in the image. The scene objects are categorized into three groups based on the cosine similarity measured by a pre-trained CLIP model~\cite{radford2021learning} between the instruction and object names, following these matching rules: (1) Target Objects (TOs): Objects explicitly mentioned in the instruction and directly involved in the action.
(2) Surrounding Objects (SOs): Foreground objects not in TOs, providing spatial and contextual cues.
(3) Background Objects (BOs): Static environmental elements ensuring scene consistency but not directly affecting the task.


Then, we encode the grouped objects using the CLIP model. The text encoder \( H \) maps object names from TOs, SOs, and BOs into textual embeddings $H(\text{TOs}) \in \mathbb{R}^{t \times D}$, $H(\text{BOs}) \in \mathbb{R}^{t \times D}$, and $H(\text{SOs}) \in \mathbb{R}^{t \times D}$. Meanwhile, the vision encoder \( P \) processes the entire image into $P(X) \in \mathbb{R}^{t \times D}$.
We assign image tokens to object groups by computing the cosine similarity between token and text embeddings, selecting the top eight tokens per group. To enhance intra-group feature representation, each group is processed through an MoE module, producing refined image token embeddings. The enhanced tokens are then concatenated with the remaining tokens for subsequent processing. This IOSP module enables the model to focus on critical objects, improving task-specific manipulation.


\subsection{Dual Gating Mixture-of-Experts }
\label{sec:method:DG-MoE}

In this section, we introduce Dual Gating Mixture-of-Expert~(DGMoE) mechanism that supports adaptive knowledge routing.
Different from vanilla MoE, which assigns a fixed number of experts to each token, our DGMoE can automatically determine the acceptance or rejection of the token assignments under the Self-Aware Expert.

As shown in Figure \ref{fig:overview}, each DGMoE layer is comprised of $K$ self-aware experts $\mathbb{E} = \{ E_1, E_2, \dots, E_K \}$.
To enable the automatic token-driven expert selection, we introduce two gating mechanisms: Token-side gating mechanism $G_t$ and Expert-side gating mechanism $G_e$.
(1) For Token-side gating, $G_t$ serves as the gating module on the token side, comprised of a soft router module and a gating residual module.
The soft router first estimates the selection scores for the hidden inputs, indicating the token preference for different experts.
Then the estimated scores from previous layer are incorporated with the scores of current layer, allowing the tokens to inherit the selection prior of the most suitable experts from previous layers.
(2) For Expert-side gating, $G_e$ can autonomously accept or reject tokens based on their acceptance threshold.
The selection scores predicted by $G_t$ is filtered with a learnable threshold, and only the experts with the most relevant tokens are activated.

Given an input token \( x \), the token-side gate \( G_t \) first computes the scores at the \( j \)-th layer by combining the soft router output and the gating residual module:
\begin{equation}
G_t^j (x) =
\begin{cases}
    W_t^j x, & \text{if } j = 1, \\
    W_t^j x + W_t^g G_t^{(j-1)}(x), & \text{otherwise},
\end{cases}
\end{equation}
where \( W_t^j \in \mathbb{R}^{D} \) is the trainable weight matrix for soft router, and \( W_t^g \in \mathbb{R}^{N \times N} \) is a trainable transformation matrix designed to aggregate the scores from the previous layer and the current layer, ensuring smooth expertise knowledge transitions across layers. Then, the token-to-expert selection scores \( s_t(x) \) are obtained by normalizing \( G_t(x) \) with a softmax function, which determines the probability distribution over the available experts.

On the self-aware expert side, the expert-side gate \( G_e \) evaluates the token scores and determines the final expert activation scores \( s_e(x) \) as follows:
\begin{equation}
s_e(x) = \text{sign}(s_t(x) - \lambda W_e),
\end{equation}
where \( W_e \) is a set of trainable self-aware parameters corresponding to the gating values of each expert, and \( \lambda \) is a scaling factor that modulates the gating threshold. In our setting, we set 
$\lambda =0.5$ to balance expert selection, ensuring neither excessive expert activation nor overly strict filtering of tokens.

Then, combining the dual gating mechanism of token-side and exper-side gating, the expert selection function is defined as follows:
\begin{equation}
g(x) =
\begin{cases}
    s_t(x), & \text{if } s_e(x) > 0, \\
    0, & \text{if } s_e(x) \leq 0,
\end{cases}
\end{equation}
where the final output token \( y \) of the DGMoE layer can be computed as the weighted sum of the selected expert outputs:
\vspace{-3mm}
\begin{equation}
y = \sum_{i=1}^{K} g_i(x) E_i(x).
\end{equation}

With the help of self-aware expert gating mechanism, our DGMoE can dynamically activate the most reclevant experts, reducing computational overhead while maintaining task-relevant feature learning.

\subsection{Expert-Driven Aggregation}
\label{sec:method:EOA}

To mitigate performance degradation caused by task heterogeneity, we propose an Expert-Driven Aggregation (EDA) strategy, which aggregates model parameters based on the similarity of expert selections across clients. And as shown in Figure~\ref{fig:overview}, we selectively aggregates only the trunk module, while the stem and head modules remain personalized and are excluded from aggregation. 
%

Given a trunk module with \( L \) layers, where each layer contains \( K \) experts, each client \( C_i \) records an expert selection matrix \( \mathbf{V}_i \in \mathbb{R}^{L \times K} \) during each round through DGMoE, where each element \( V_{i}^{(l,k)} \) represents the number of times expert \( k \) is activated in the \( l \)-th layer. We extract the expert selection vector for layer \( l \) of client \( C_i \) as:
\begin{equation}
\mathbf{v}_i^{(l)} = ( V_{i}^{(l,1)}, V_{i}^{(l,2)}, \dots, V_{i}^{(l,K)} ).
\end{equation}

Then we compute the pairwise similarity between clients based on expert selection vectors. 
The similarity between clients \( C_i \) and \( C_j \) at layer \( l \) is defined as follows:
\begin{equation}
s_{i,j}^{(l)} = \frac{\mathbf{v}_i^{(l)} \cdot \mathbf{v}_j^{(l)}}{\|\mathbf{v}_i^{(l)}\| \|\mathbf{v}_j^{(l)}\| }.
\end{equation}

Using this similarity, we determine the aggregation weight \( w_{l,i} \) for each client \( C_i \) at layer \( l \) as follows:
\begin{equation}
w_{l,i} = \frac{\sum_{C_j \in C} s_{i,j}^{(l)}}{\sum_{C_j \in C} \sum_{C_i \in C} s_{i,j}^{(l)} }.
\end{equation}

This weighting strategy ensures that clients with higher expert selection similarity contribute more to each other's model updates, allowing for more effective and task-aligned aggregation across task-heterogeneous clients.

\subsection{Algorithms}
\bfsection{Clients} The overall workflow of our \name{} is shown in ~\cref{alg:clients} and ~\cref{alg:server}.~At the start of each training round $t$, each client processes task-specific data using Instruction-Oriented Scene-Parsing to extract structured features, followed by Dual Gating MoE  for adaptive expert selection.The updated trunk parameters and expert selection counts are then transmitted to the server for global aggregation. The detailed client-side process is described in Algorithm~\cref{alg:clients}.

\bfsection{Server} On the server side, the server receives expert selection statistics and trunk updates from all participating clients and performs Expert-Driven Aggregation, which dynamically assigns aggregation weights based on expert selection similarities. The aggregated global trunk module is then redistributed to clients for the next training round.
The process is elaborated in \cref{alg:server}.

\begin{algorithm}[!t]
\caption{\textbf{FedVLA: Clients}}
\label{alg:clients}
\textbf{Input}: Each client $C_i$ has a local task $T_i$ and dataset $D_i = \{o_i, l_i, s_i, a_i\}$, along with local parameters $\phi_i$, $w_i$, and $\theta_i$ corresponding to the stem, trunk, and head modules, respectively. \\
\begin{algorithmic}[1]
\STATE \textbf{Local\_Train($C_i, \phi_i, \theta_i$):}
\FOR{each local epoch $e$ from $1$ to $E_c$}
    \FOR{each batch $(o_i, l_i, s_i, a_i)$ sampled from $D_i$}
    \vspace{4pt}
    \STATE // Step 1: Instruction-Oriented Scene-Parsing
    \vspace{0pt}
    \STATE \texttt{TOs}, \texttt{SOs}, \texttt{BOs} $\leftarrow \{o_i, l_i\}$
    \quad 
    \STATE $F_{TOs}, F_{\mathrm{SOs}}, F_{\mathrm{BOs}} \gets \text{CLIP}(\texttt{TOs}, \texttt{SOs}, \texttt{BOs})$
    \vspace{1pt}
    \STATE $F_O \leftarrow \mathrm{CLIP}(o_i)$
    \vspace{1pt}
    \STATE $X_{\mathrm{TO}} \leftarrow \mathrm{topk}(\mathrm{cosine}(F_{\mathrm{TOs}}, F_O), F_O)$
    \vspace{1pt}
    \STATE $X_{\mathrm{SO}} \leftarrow \mathrm{topk}(\mathrm{cosine}(F_{\mathrm{SOs}}, F_O), F_O)$
    \vspace{1pt}
    \STATE $X_{\mathrm{BO}} \leftarrow \mathrm{topk}(\mathrm{cosine}(F_{\mathrm{BOs}}, F_O), F_O)$
    \vspace{1pt}
    \STATE $X_{\mathrm{in}} = \mathrm{concat}(X_{\mathrm{TO}},\,X_{\mathrm{SO}},\,X_{\mathrm{BO}},\,s_i)$
    \vspace{4pt}
    \STATE // Step 2: Forward pass with DGMoE
    \vspace{0pt}
    \STATE Let $x \in MHA(X_{\mathrm{in}})$ \\
    \vspace{1pt}
          $s_t(x) = \mathrm{softmax}\bigl(W_t(x)\bigr)$ 
    \quad \quad // token-gate
    \vspace{1pt}
    \STATE $s_e(x) = \mathrm{sign}(s_t(x) - \lambda\,W_e)$ 
    \quad // expert-gate
    \vspace{1pt}
    \STATE $g(x) = s_t(x) \cdot \mathbb{I}(s_e(x) > 0)$ \quad // expert activation
    \vspace{-9pt}
    \STATE $\mathbf{v}^{(l)},y = \displaystyle\sum_{k=1}^{K} g(x)\,E_k(x)$ 
    \quad // final output
    \vspace{4pt}
    \STATE // Step 3: Model parameter updates
    \vspace{0pt}
    \STATE $\mathcal{L}_i = \mathrm{Loss}(f(\phi_i,\,\theta_i,\,w_i),\,a_i)$
    \quad 
    \vspace{1pt}
    \STATE $(\phi_i,\,\theta_i,\,w_i)\!\leftarrow\!(\phi_i,\,\theta_i,\,w_i)\!-\!\eta\,\nabla\,\mathcal{L}_i$
    \quad 
    \ENDFOR
\ENDFOR

\STATE \textbf{return} $\theta,\mathbf{v}^{(l)}$ to the server.
\end{algorithmic}
\end{algorithm}

\begin{algorithm}[!t]
\caption{\textbf{FedVLA: Server}}
\label{alg:server}
\textbf{Input}: $T$ is the number of training rounds. $\theta$ represents the global trunk parameters. Each client provides $\mathbf{v}^{(l)}$, the expert selection counts vector for each layer $l$.\\

\begin{algorithmic}[1]
\STATE \textbf{Server\_Execute:}
\FOR{each round $t$ from $1$ to $T$}
    \STATE // Expert-Driven Aggregation
    \FOR{each trunk layer $l$}
        \FOR{each pair of clients $C_i, C_j$}
            \STATE  $s_{i,j}^{(l)} = \frac{ \mathbf{v}_i^{(l)} \cdot \mathbf{v}_j^{(l)} }{ \| \mathbf{v}_i^{(l)} \| \| \mathbf{v}_j^{(l)} \| }$
            // expert selection similarity
            \vspace{2pt}
            \STATE  $w^{(l,i)} = \frac{\sum_{j \in C} s_{i,j}^{(l)}}{\sum_{j \in C} \sum_{i \in C} s_{i,j}^{(l)}}$
            // aggregation weight
            \vspace{2pt}
        \ENDFOR
        \STATE  $\theta^{(l)} \gets \sum_{i=1}^{N} w^{(l,i)} \cdot \theta^{(l,i)}$
        \quad \textit{// update $\theta$}
    \ENDFOR
    \STATE Broadcast updated $\theta$ to all clients.
\ENDFOR
\end{algorithmic}
\end{algorithm}
\section{Experiments}
\label{sec:exp}



\noindent\textbf{Implementation Details.}  
We validate our \name{} in both simulation and real-world scenarios, where each client performs a single task, resulting in a non-iid data distribution. We employ the pretrained HPT~\cite{wang2025scaling} as the backbone of our VLA model and train it for 1,000 communication rounds between clients and the server, with each client performing local training for 5 epochs per round.
In simulation, the model is trained with a learning rate of $5\times10^{-6}$, while in real-world settings, the learning rate is set to $2\times10^{-5}$. Each client trains locally with a batch size of 256 using the Adam optimizer. The experiments are conducted on a workstation equipped with 2 GeForce GTX 4090Ti GPUs, 48 Intel Xeon CPUs, and 128GB of memory. 

\noindent\textbf{Evaluation.} For evaluation, the success and failure of a trial are recoreded as 1 and 0. Each task takes for 15 trials and repeats for 5 times. 
Since this is the first work on federated VLA learning, we compare \name{} with two methods. The centralized training (Centralized) aggregates all data on a single server for joint training, serving as an upper bound in performance. The federated averaging (FedAvg)~\cite{mcmahan2017communication} is a strong FL baseline where models are trained locally on each client and simply averaged on the server.


\subsection{Simulation}
\noindent\textbf{Environment.} We utilize the MuJoCo engine~\cite{mandlekar2021matters} to simulate tasks from the Meta-World robot manipulation benchmark~\cite{yu2020meta}. All tasks are performed using a Sawyer robot with a parallel gripper, positioned at the center of a dinner table. Visual data is captured by three fixed RGB cameras with a resolution of $128\times128$, providing corner, top, and shoulder views. For collision detection and dynamics simulation, we employ  official physics engines to ensure accurate robotic interactions within the simulation environment.

\noindent \textbf{Tasks.} 
We evaluate our \name{} on four household tasks: Door Lock, Close Drawer, Sweep Into, and Open Window. For each task, we collect approximately 30–80 episodes, each consisting of 40–100 steps. The initial pose of robot and environment layouts are randomized to assess the generalization ability of \name{}.

\begin{table}[h]
    \centering
    \resizebox{0.47\textwidth}{!}{
    \begin{tabular}{lccccc}
        \toprule
        \multirow{2}{*}{\textbf{Method}} & \textbf{Avg.} & \textbf{Door} & \textbf{Close} & \textbf{Sweep} & \textbf{Open} \\
        & \textbf{Success} & \textbf{Lock} & \textbf{Drawer} & \textbf{Into} & \textbf{Window} \\
        \midrule
        Centralized  & \textbf{65.0} & \textbf{86.7}  & 73.3 & 53.3 & \textbf{46.7} \\
        FedAvg       & 51.7 & 66.7 & 73.3 & 40.0 & 26.7 \\
        FedVLA (Ours)& 63.3 & 80.0 & \textbf{80.0} & \textbf{53.3} & 40.0 \\
        \bottomrule
    \end{tabular}
    }
    \caption{Simulation evaluation across four tasks.}
    \label{tab:simu_success}
\end{table}


\noindent \textbf{Performance Comparison.} The evaluation results in the simulation environment are presented in Table~\ref{tab:simu_success}. \name{} achieves an average success rate of 63.3\%, closely approaching the 65.0\% of centralized training. Notably, in the Close Drawer task, \name{} surpasses centralized training by a success rate margin of about 7.0\%. Furthermore, \name{} consistently outperforms FedAvg, which only achieves an average success rate of 51.7\%. This performance gap can be attributed to FedAvg overlooking task heterogeneity in federated learning. In contrast, \name{} explicitly accounts for task heterogeneity during aggregation by leveraging expert-driven model updates, ensuring that knowledge from diverse clients is effectively integrated to enhance generalization across different tasks.

\subsection{Real-World}
\noindent \textbf{Hardware.} For real-world experiments, we employ a UR3 robotic arm with six degrees of freedom, equipped with a one-DoF parallel gripper. A RealSense D435i RGB-D camera is mounted above the robot, capturing RGB images in real time at a resolution of $1280\times720$ with a frame rate of approximately 30 Hz. To simulate real-world variations, we introduce moderate randomness in object placement and lighting conditions to better reflect practical deployment scenarios.

\begin{figure}[!h]
    \centering
    \includegraphics[width=0.47\textwidth]{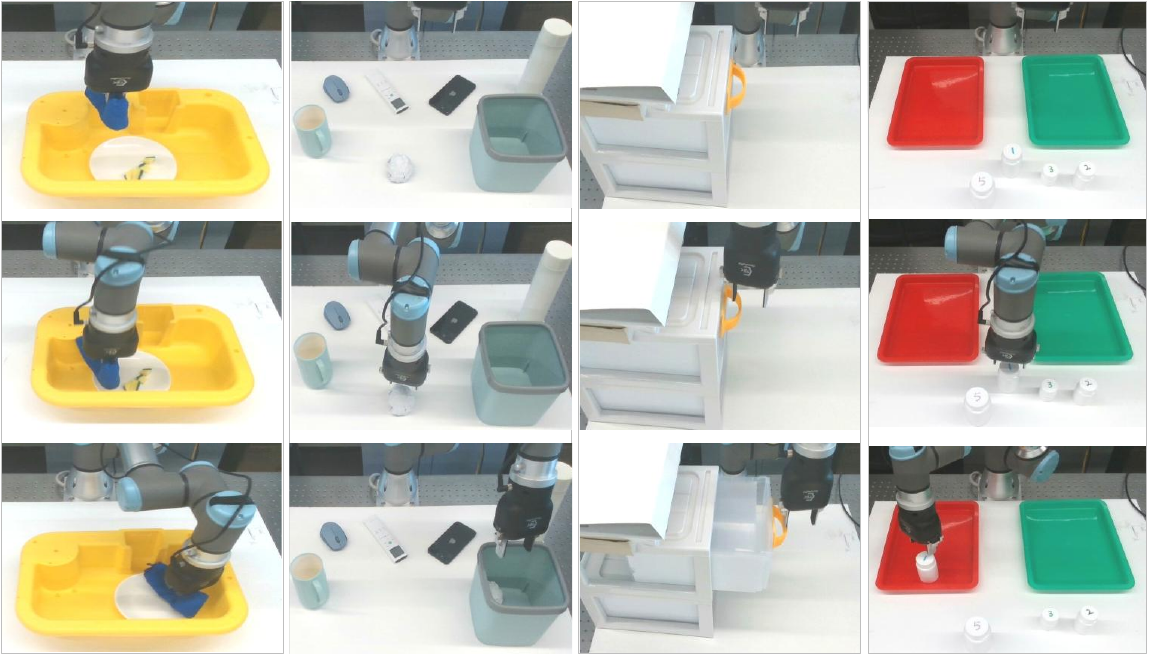}
   \caption{Real-World Tasks. Each column represents a different task, while each row demonstrates keyframes of task execution. From left to right, the tasks are: \textit{Clean Up}, where the robot wipes the trash from plate using a cloth. \textit{Trash Collection}, in which it throws crumpled paper into a trash bin. \textit{Open Drawer}, which requires grasping and pulling a drawer open, and \textit{Sorting Pills}, where pill bottles are placed into trays of designated color.}
    \label{fig:realtasks}
\end{figure}

\noindent\textbf{Tasks.} We collect real-world robotic demonstrations for household-related tasks, including Clean Up, Trash Collection, Open Drawer, and Sorting Pills, as shown in Figure~\ref{fig:realtasks}. For data collection, we employ an expert teaching method, using inverse kinematics to compute the robot poses corresponding to key point data while recording RGB images and pose data throughout the trajectory. Each task includes approximately 50 demonstrations, with each trajectory consisting of 20 to 80 steps.

\noindent\textbf{Performance Comparison.} The real-world experimental results are summarized in Table~\ref{tab:real_success}. \name{} achieves an average task success rate of 63.3\%, closely matching the 63.4\% of centralized training. Notably, it attains the same success rates as the centralized model in complex tasks such as Trash Collection (46.7\%) and Sorting Pills (73.3\%). In contrast, FedAvg exhibits a significant performance drop across all tasks, with an average success rate of 53.3\%, which is 10.0\% lower than our method. These results indicate that \name{} effectively bridges the performance gap between federated and centralized training. While FedAvg merely averages model parameters across clients, it struggles to generalize effectively in VLA learning, where multimodal information—including textual instructions, scene images, and execution trajectories—is involved, requiring strong feature extraction capabilities. The proposed Dual Gating MoE and Expert-Driven Aggregation mechanisms enable the model to adaptively learn diverse patterns for robotic manipulation, enhancing its ability to handle complex tasks in federated settings.

\begin{table}[h]
    \centering
    \resizebox{0.47\textwidth}{!}{ 
    \begin{tabular}{lccccc}
       \toprule
        \multirow{2}{*}{\textbf{Method}} & \textbf{Avg.} & \textbf{Clean} & \textbf{Trash} & \textbf{Open} & \textbf{Sorting} \\
        & \textbf{Success} & \textbf{Up} & \textbf{Collection} & \textbf{Drawer} & \textbf{Pills} \\
        \midrule
        Centralized  & \textbf{63.4} & 46.7 & 46.7 & \textbf{86.7} & 73.3 \\
        FedAvg       & 53.3 & 46.7 & 40.0 & 60.0 & 66.7 \\
        FedVLA (Ours)& 63.3 & \textbf{53.3} & \textbf{46.7} & 80.0 & \textbf{73.3} \\
        \bottomrule
    \end{tabular}
    }
    \caption{Real-world evaluation across four tasks.}
    \label{tab:real_success}
\end{table}

\subsection{Ablation Studies}

To further explore the effectiveness of the IOSP, DGMOE and EDA  in FedVLA, we conduct ablation experiments by individually removing each module while keeping the other components.
Specifically, the experimental setups are as follows: (1) remove the IOSP module and directly input the entire scene image and proprioception data into the encoder (w/o IOSP), (2) replace the DGMoE layers with standard FFN, eliminating adaptive expert selection (w/o DGMOE), and (3) perform vanilla FedAvg aggregation instead of EDA (w/o EDA).
We compare the task success rate and record the validation loss during the training process.
\vspace{-1mm}
\begin{table}[h]
    \centering
    \resizebox{0.47\textwidth}{!}{ 
     \begin{tabular}{c|ccccc}
        \toprule
        \multirow{2}{*}{\textbf{Method}} & \textbf{Avg.} & \textbf{Clean} & \textbf{Trash} & \textbf{Open} & \textbf{Sorting} \\
        & \textbf{Success} & \textbf{Up} & \textbf{Collection} & \textbf{Drawer} & \textbf{Pills} \\
        \midrule
        w/o IOSP  & 41.1 & 40.0 & 13.3 & 66.7 & 46.7 \\
        w/o DGMoE & 31.7 & 20.0 & 20.0 & 46.7 & 40.0 \\
        w/o EDA  & 26.7 & 26.7 & 20.0 & 33.3 & 26.7 \\
        FedVLA (Ours)  & \textbf{63.3} & \textbf{53.3} & \textbf{46.7} &  \textbf{80.0} & \textbf{73.3} \\
        \bottomrule
    \end{tabular}
    }
    \caption{Ablation studies of proposed FedVLA without IOSP, DGMoE, and EDA.}
    \label{tab:realworld_ablation}
\end{table}
\vspace{-3mm}

From Table~\ref{tab:realworld_ablation}, we can see that each module contributes significantly to the performance of FedVLA, and the lack of each module will lead to a decrease in the performance of the entire system.
Specifically, removing IOSP leads to the highest performance drop in the Trash Collection task from 46.7\% to 13.3\%, because the number of objects in Trash Collection task is more than the others.
The IOSP module discriminates each object from the other for taking the right action to adapt to multi-objects scenarios. 
DGMoE module strongly relates to the Clean Up task, leading to a success rate drop from 53.3\% to 20.0\%.
The Clean Up task is harder in the manipulation aspect as it requires the robot interaction for multiple steps.
The FedVLA depends on the DGMoE module to adaptively select the appropriate number of experts based on different types of tokens, enabling collaborative learning for interaction-rich tasks. 
Furthermore, the removal of EDA gives the largest reduction in average success rate from 63.3\% down to 26.7\%
As a common aggregation cancels out the knowledge of the experts in the DGMoE module.

\begin{figure*}[h]
    \centering
    \begin{subfigure}{0.24\textwidth}
        \centering
        \includegraphics[width=\textwidth]{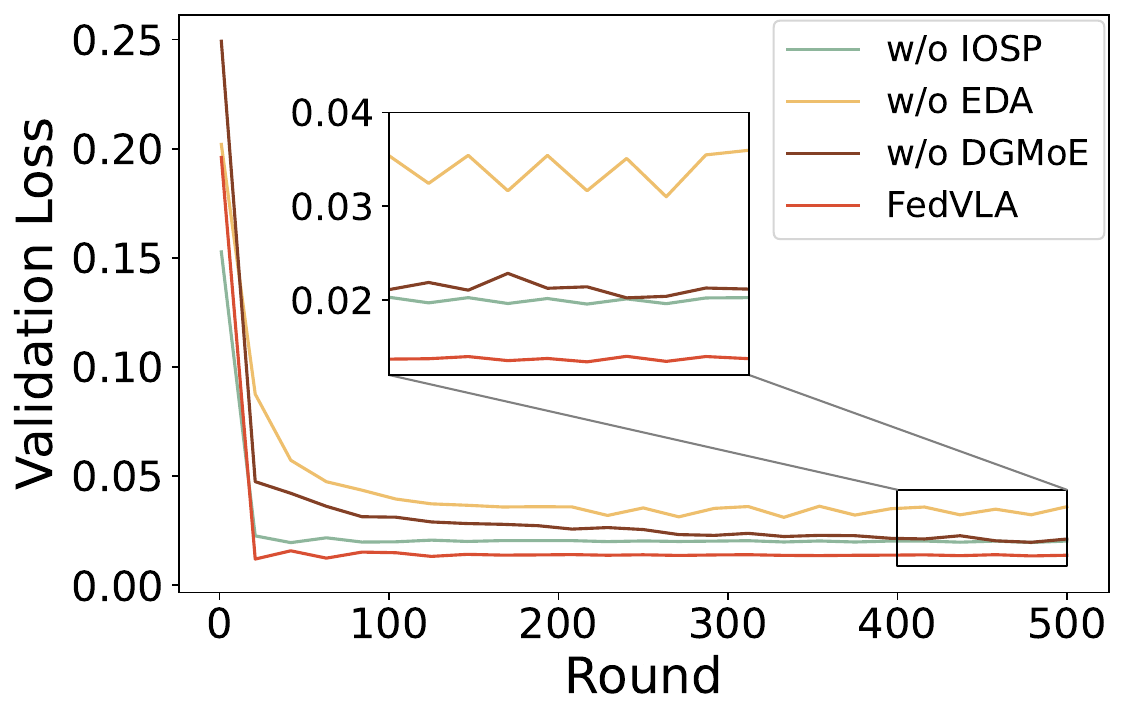}
        \caption{Open Drawer}
        \label{fig:valoss1}
    \end{subfigure}
    \begin{subfigure}{0.24\textwidth}
        \centering
        \includegraphics[width=\textwidth]{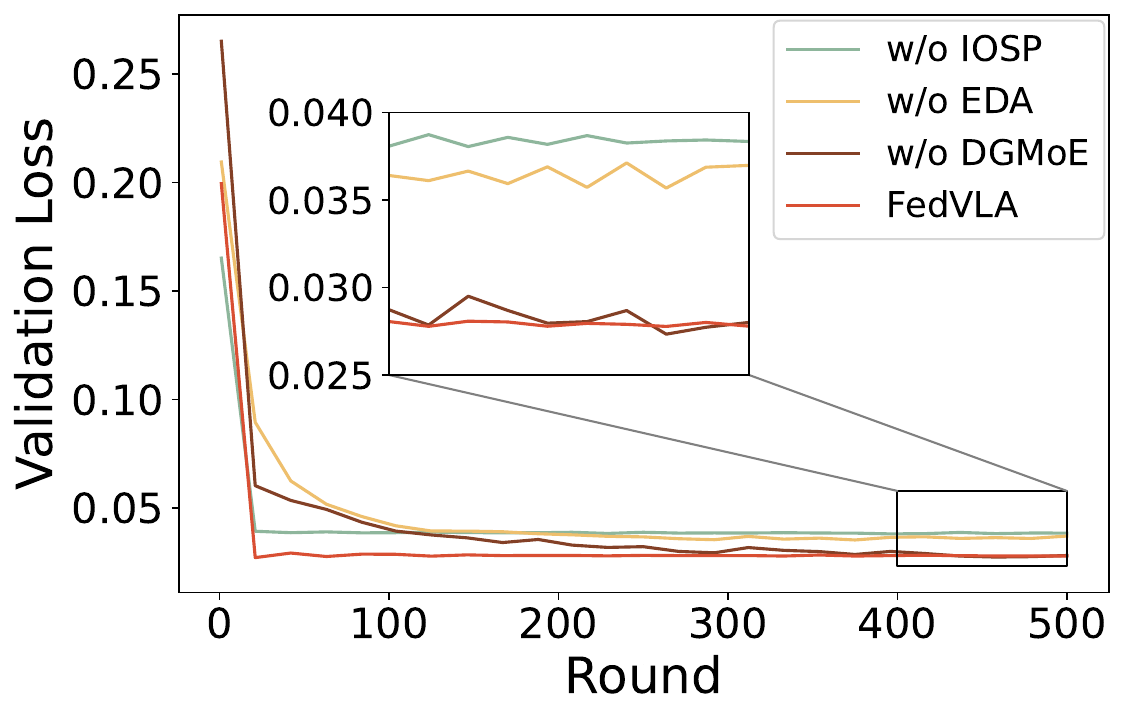}
        \caption{Sorting Pills}
        \label{fig:valoss2}
    \end{subfigure}
    \begin{subfigure}{0.24\textwidth}
        \centering
        \includegraphics[width=\textwidth]{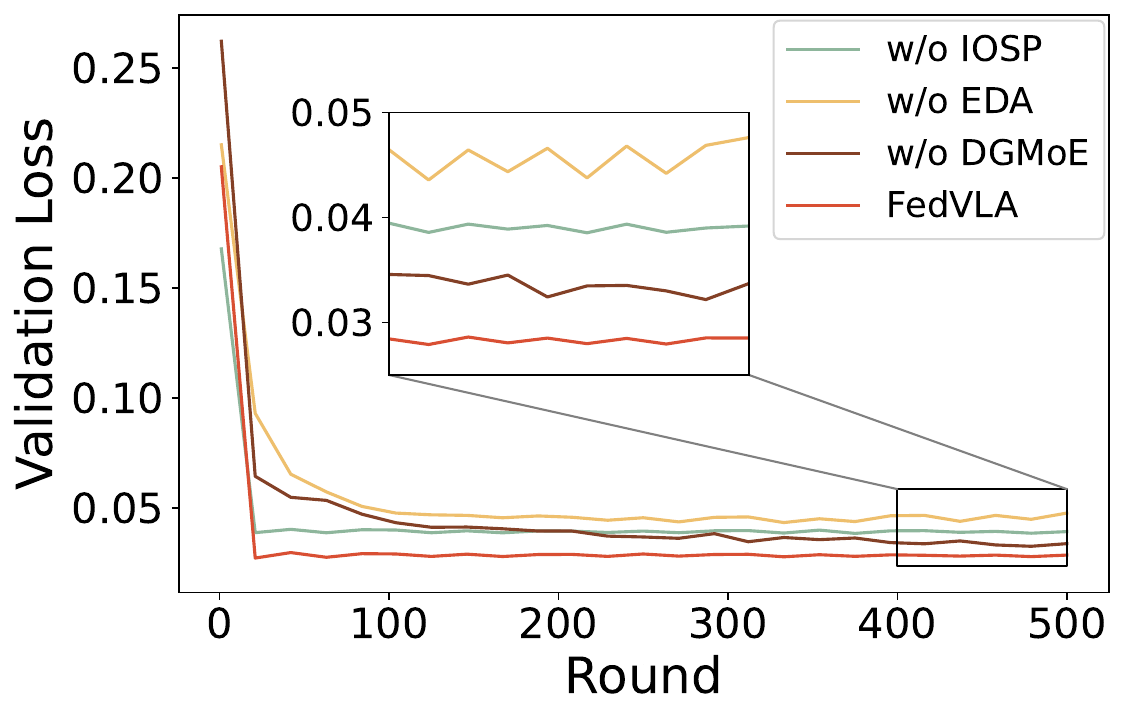}
        \caption{Trash Collection}
        \label{fig:valoss3}
    \end{subfigure}
    \begin{subfigure}{0.24\textwidth}
        \centering
        \includegraphics[width=\textwidth]{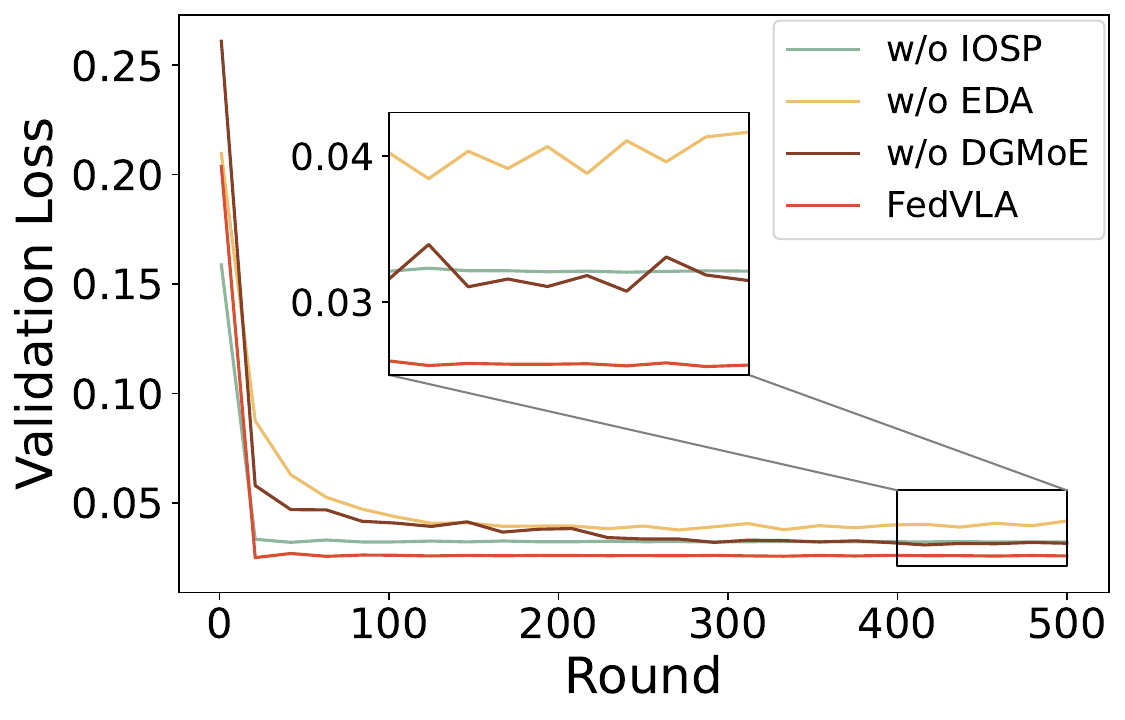}
        \caption{Clean Up}
        \label{fig:valoss4}
    \end{subfigure}
    
    \caption{Validation loss comparison in ablation study across four tasks. The x-axis represents the training rounds, while the y-axis represents the validation loss. Each subplot includes a zoomed-in inset focusing on the validation loss fluctuations between rounds 400 and 500, illustrating the fine-grained differences in convergence trends. }
    \label{fig:valoss}
\end{figure*}

Figure~\ref{fig:valoss} presents the validation loss of the ablation experiments during the first 500 training epochs.
We can see that FedVLA consistently achieves the fastest convergence speed, along with the lowest and most stable validation losses across all tasks. 
Specifically, we can discover that (1) without IOSP causes validation losses notably increase in the multi-object scenarios (depicted in Figure~\ref{fig:valoss2} and Figure~\ref{fig:valoss3}), (2) the removal of DGMoE results in significantly increased and fluctuating losses the interaction-rich tasks, (3) removing the EDA module causes relatively higher validation losses in most tasks. These findings strongly support that IOSP helps decompose complex visual scenes into task-relevant object representations, DGMoE module routes the adaptive knowledge, EDA ensures the integrity and effectiveness of the aggregation. The integration of these modules together results in a architecture that supports FedVLA’s robustness and adaptability across diverse tasks.

\begin{figure*}[htpb]
    \centering
    \begin{subfigure}{0.24\textwidth}
        \centering
        \includegraphics[width=\linewidth]{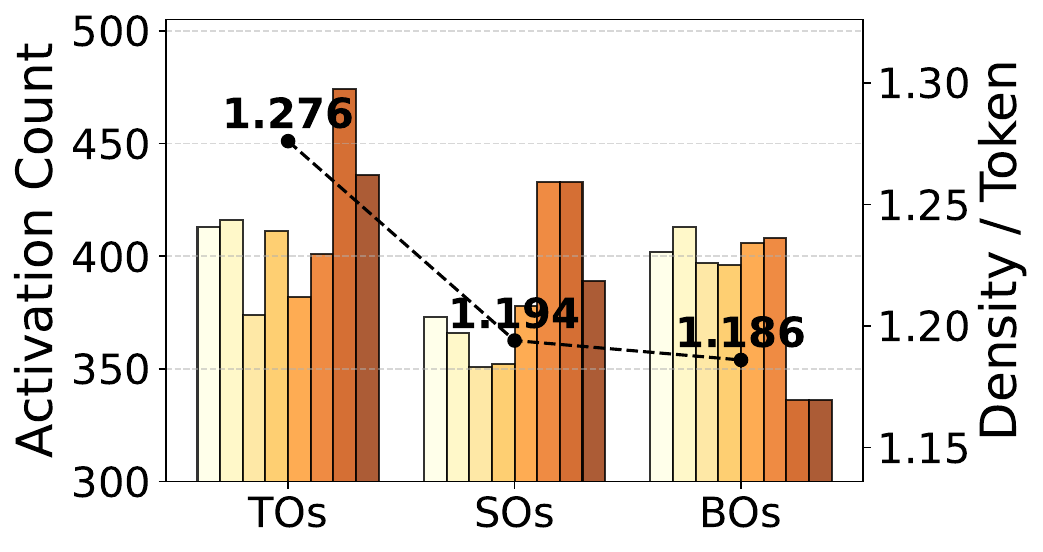}
        \caption{Open Drawer}
        \label{fig:expert_object1}
    \end{subfigure}
    \hfill
    \begin{subfigure}{0.24\textwidth}
        \centering
        \includegraphics[width=\linewidth]{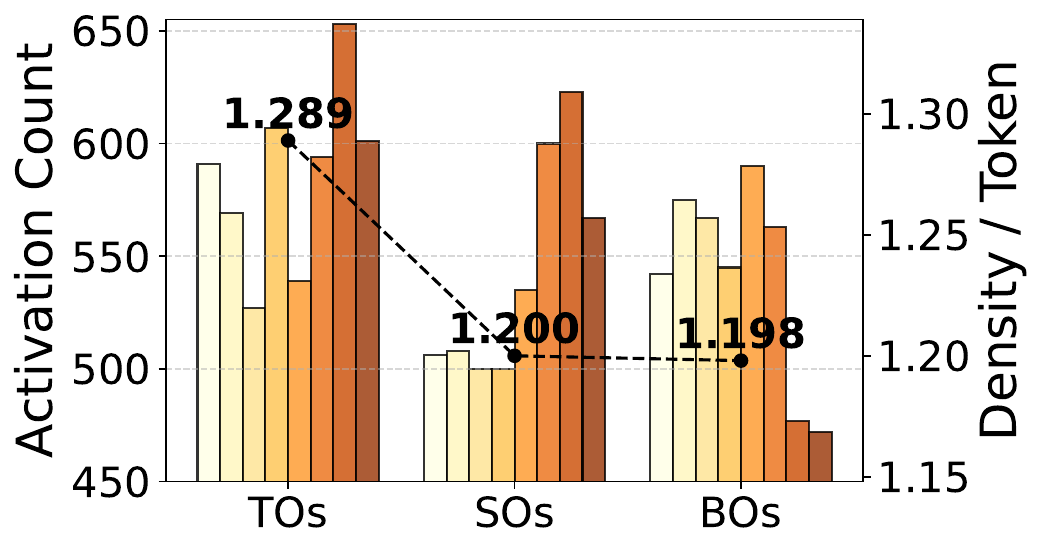}
        \caption{Sorting Pills}
        \label{fig:expert_object2}
    \end{subfigure}
    \hfill
    \begin{subfigure}{0.24\textwidth}
        \centering
        \includegraphics[width=\linewidth]{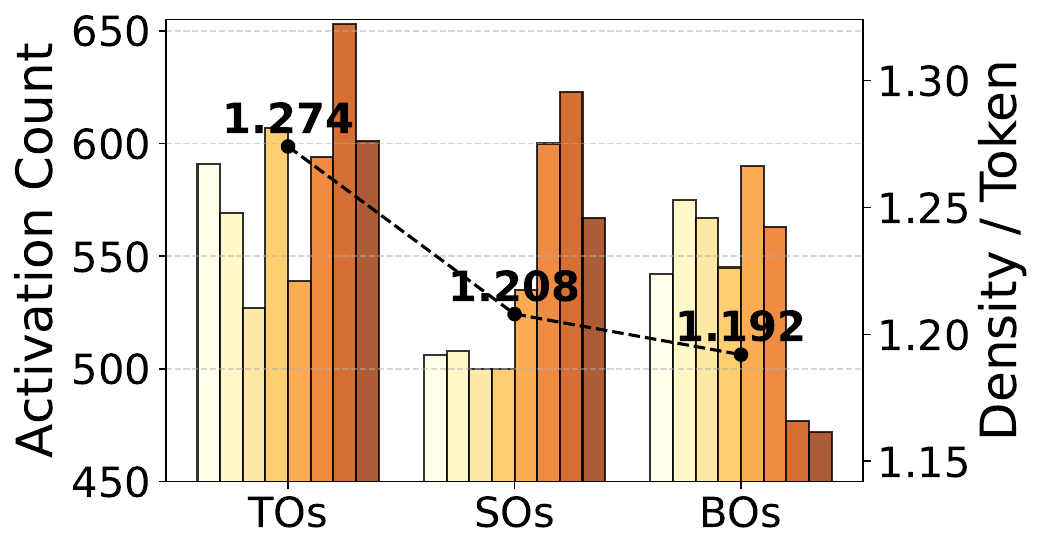}
        \caption{Trash Collection}
        \label{fig:expert_object3}
    \end{subfigure}
    \hfill
    \begin{subfigure}{0.24\textwidth}
        \centering
        \includegraphics[width=\linewidth]{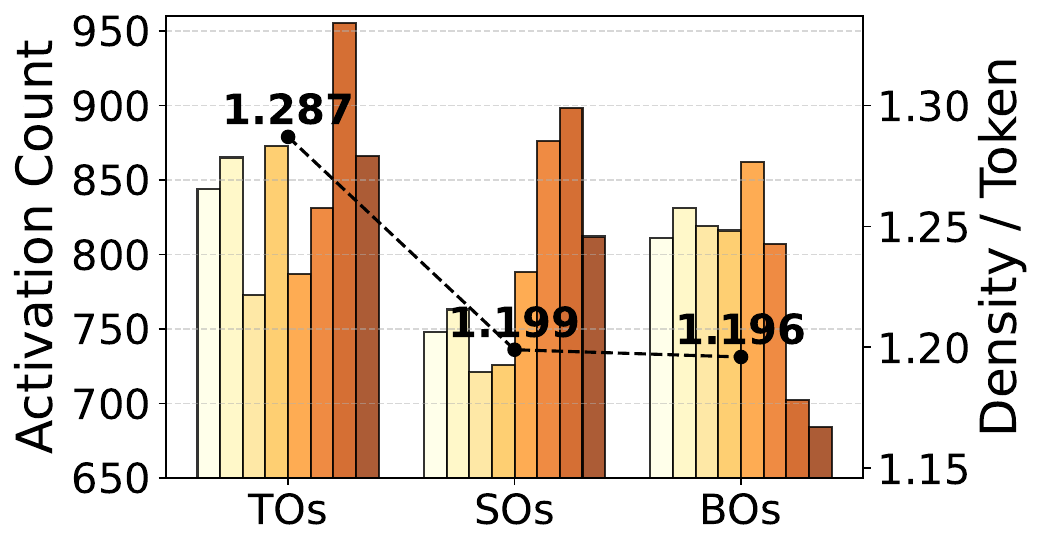}
        \caption{Clean Up}
        \label{fig:expert_object4}
    
    \end{subfigure}
    \caption{Visualization of expert activation distribution per task for different object types. The x-axis represents the three object groups (target objects, surrounding objects, and background objects), where each group contains eight sub-bars corresponding to different experts selected. The left y-axis indicates the activation count, showing how frequently experts are selected for each object type. The right y-axis represents the density of experts per token, illustrating the average number of experts assigned per token in each object category. The dashed line represents the average expert density per token across different object groups, highlighting the dynamic expert selection mechanism of DGMoE.}
    \label{fig:expert_selection}
\end{figure*}

\subsection{Analysis on DGMoE}
The DGMoE is designed to enhance model efficiency by selectively activating experts based on task complexity and their corresponding object types. 
In contrast, supporting objects and background objects, which provide contextual spatial information, activate fewer experts compared to target objects. 
Notably, target objects tend to activate experts 4, 7, and 8 more frequently, supporting objects prefer to select experts 6, 7, and 8, whereas background objects favor experts 2, 3, and 5. 
This demonstrates that DGMoE not only dynamically allocates computational resources based on object relevance but also selectively routes information to specialized experts, enhancing efficiency and adaptability across different object categories.

Furthermore, we quantitatively measure the average expert density per token across different tasks, also as shown in Figure~\ref{fig:expert_selection}. 
The proposed DGMoE achieves consistently lower expert activation densities compared to vanilla top-$k$ MoE, which assigns tokens to a fixed number of experts, where 
$k$ is typically an integer greater than one. The average number of activated experts per token is 1.219, 1.229, 1.225, and 1.227 for Open Drawer, Sorting Pills, Trash Collection, and Clean Up, respectively. Such dynamic and sparse activation directly translates into significantly reduced forward-pass computation, making DGMoE especially suitable for deployment on resource-constrained edge devices.

\section{Conclusion}
In this work, we introduced \name{}, a novel federated vision-language action learning framework that preserves privacy and effectively balances data privacy and task performance. To address the challenges of task heterogeneity and efficient knowledge aggregation in decentralized settings, we proposed three key components: the Instruction-Oriented Scene-Parsing module, which enhances feature extraction by decomposing raw observations into structured object-level representations guided by task instructions. The Dual Gating Mixture-of-Experts, which introduces self-aware experts capable of bidirectional token selection, optimizing computational efficiency while preserving task performance, and the Expert-Driven Aggregation strategy, which dynamically assigns aggregation weights based on expert similarity across clients. Extensive experiments in both simulated and real-world environments demonstrate that \name{} achieves performance comparable to centralized training while ensuring privacy protection. 
We believe these results may contribute to the development of more generalizable and scalable federated learning frameworks for robotic manipulation.

\vspace{1em}
\bfsection{Acknowledgements} This research was supported by Guangdong Basic and Applied Basic Research Foundation (No. 2024A1515011774).

\label{sec:con}
{
    \small
    \bibliographystyle{ieeenat_fullname}
    \bibliography{main}
}

\end{document}